\documentclass[letterpaper]{article} 
\usepackage{aaai2025}  
\usepackage{times}  
\usepackage{helvet}  
\usepackage{courier}  
\usepackage[hyphens]{url}  
\usepackage{graphicx} 
\usepackage{natbib}  
\usepackage{caption} 
\usepackage{algorithm}
\usepackage{algorithmic}

\usepackage{multirow}
\usepackage{amssymb}

\usepackage{makecell}

\usepackage{newfloat}
\usepackage{listings}
\DeclareCaptionStyle{ruled}{labelfont=normalfont,labelsep=colon,strut=off} 
\floatstyle{ruled}
\newfloat{listing}{tb}{lst}{}
\floatname{listing}{Listing}
\title{Exploring 3D Reasoning-Driven Planning: From Implicit Human Intentions to Route-Aware Activity Planning}

\author{
    Xueying Jiang\textsuperscript{\rm 1},
    Wenhao Li\textsuperscript{\rm 1},
    Xiaoqin Zhang\textsuperscript{\rm 2},
    Ling Shao\textsuperscript{\rm 3},
    Shijian Lu\textsuperscript{\rm 1}\thanks{Corresponding author.}
}
\affiliations{
    \textsuperscript{\rm 1}College of Computing and Data Science, Nanyang Technological University, Singapore\\
    \textsuperscript{\rm 2}College of Computer Science and Technology, Zhejiang University of Technology, China\\
    \textsuperscript{\rm 3}UCAS-Terminus AI Lab, University of Chinese Academy of Sciences, China\\

}

\usepackage{bibentry}

\begin{document}

\maketitle

\begin{abstract}
3D task planning has attracted increasing attention in human-robot interaction and embodied AI thanks to the recent advances in multimodal learning. However, most existing studies are facing two common challenges: 1) heavy reliance on explicit instructions with little reasoning on implicit user intention; 2) negligence of inter-step route planning on robot moves. We address the above challenges by proposing 3D Reasoning-Driven Planning, a novel 3D task that reasons the intended activities from implicit instructions and decomposes them into steps with inter-step routes and planning under the guidance of fine-grained 3D object shapes and locations from scene segmentation. We tackle the new 3D task from two perspectives. First, we construct ReasonPlan3D, a large-scale benchmark that covers diverse 3D scenes with rich implicit instructions and detailed annotations for multi-step task planning, inter-step route planning, and fine-grained segmentation. Second, we design a novel framework that introduces progressive plan generation with contextual consistency across multiple steps, as well as a scene graph that is updated dynamically for capturing critical objects and their spatial relations. Extensive experiments demonstrate the effectiveness of our benchmark and framework in reasoning activities from implicit human instructions, producing accurate stepwise task plans and seamlessly integrating route planning for multi-step moves. The dataset and code will be released.
\end{abstract}

\section{Introduction}
With the rapid development of multimodal learning, 3D task planning has become an essential component in various robot tasks that involve frequent interactions with humans and the 3D world. Traditional robot task planning relies heavily on explicit instructions. However, it is critical to empower robots to reason the intended activity from implicit human instructions since human instructions in natural language are often ambiguous and imprecise. In addition, the ability of generating a detailed task plan for the reasoned activity is critical as well for navigating 3D scenes, interacting with relevant scene objects, and accomplishing complex real-world tasks. At present, reasoning implicit human instructions, and further planning a reasonable sequence of executable steps with inter-step routes remains a grand challenge in human-robot interaction and embodied AI.

Several prior studies~\cite{wu2023embodied, shridhar2020alfred, song2023llm, chen2024grounded, hong20233d} have investigated 3D task planning, but most struggle in activity reasoning from implicit instructions and inter-step route planning. For example, ALFRED~\cite{shridhar2020alfred} achieves task planning based on explicit instructions, but it cannot accurately reason specific activities from vague or implicit human instructions. TaPA~\cite{wu2023embodied} studies decomposing human instructions into separate action steps, but they focus on task decomposition only without considering much about inter-step route planning. At present, most existing approaches are facing various challenges while addressing activity reasoning, task planning, and route planning simultaneously, struggling with comprehending users' implicit instructions and generating route plans.

\begin{figure*}[t]
    \centering
    \includegraphics[width=0.9\linewidth]{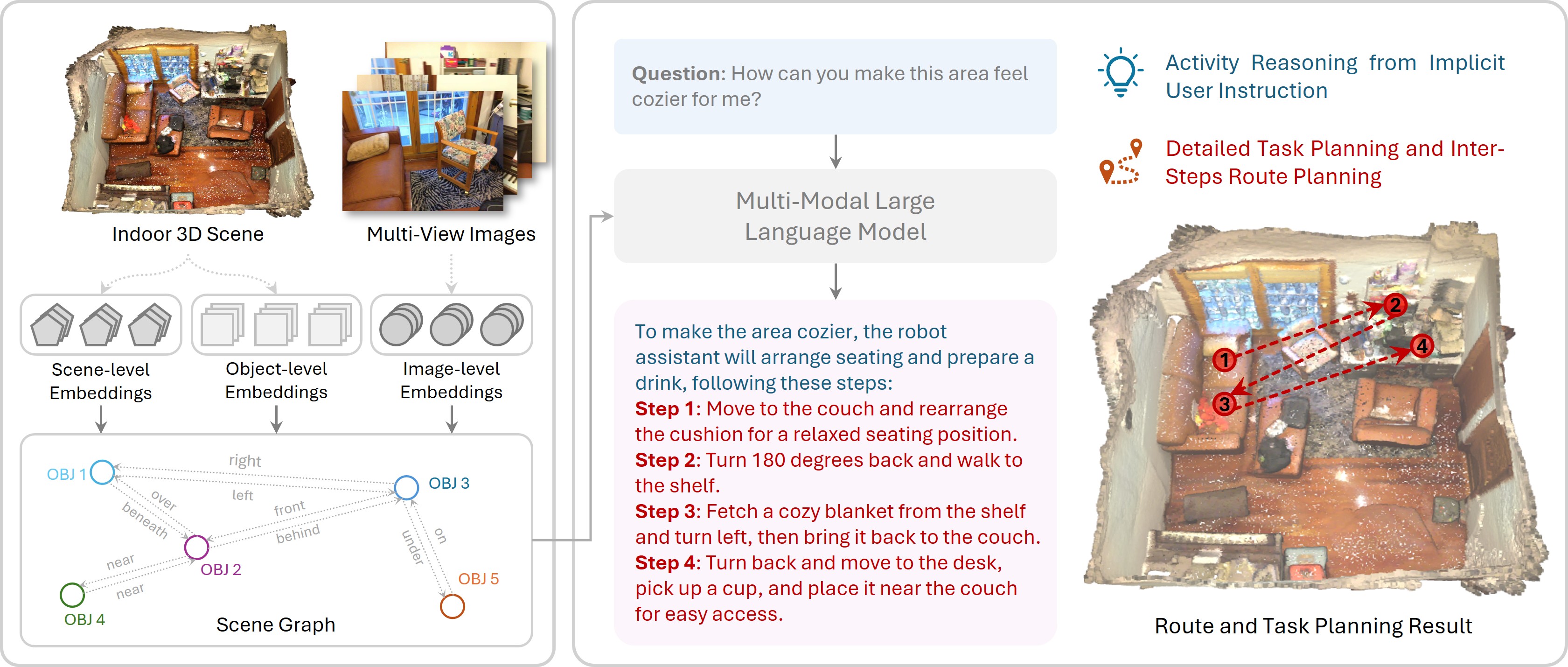}
    \caption{The proposed 3D Reasoning-Driven Planning enables reasoning activities underneath user’s implicit instructions. It can generate detailed executable steps for the reasoned activities within 3D scenes, as well as consistent inter-step route planning with object shapes and object locations derived from fine-grained 3D segmentation.}
    \label{fig:motivation}
\end{figure*}

We propose 3D Reasoning-Driven Planning, a novel 3D task that reasons the underlying intention and activities from implicit human instructions, decomposes the reasoned activities into multiple executable step-by-step plans, and performs inter-step route planning with 3D scene understanding. We tackle the new 3D task from two perspectives. First, we construct ReasonPlan3D, a large-scale and comprehensive benchmark that comprises diverse 3D scenes, various implicit human instructions and corresponding activities, detailed step-by-step plans, and annotations of the inter-step route planning. The benchmark comprises segmentation annotations of objects in scenes which benefit 3D scene understanding and concurrently provide fine-grained guidance for generating accurate route planning between steps. ReasonPlan3D features superb diversity in actions, objects, and movement types, supporting complex reasoning and planning across 3D scenarios with various activities.

In addition, we design \textbf{SHARP}, a novel 3D reasoning-driven planning framework that enables \textbf{S}cene-grap\textbf{H}-based \textbf{A}ctivity \textbf{R}easoning and \textbf{P}lanning from implicit user instructions and generates executable plans with detailed routes between steps. The framework comes with a novel progressive plan generation strategy that produces step-by-step planning by referring to historical steps, enabling contextual consistency throughout all planned steps for the reasoned activity. We construct a scene graph for each scene to model the relation among various objects in scenes, and design dynamic graph modulation (DGM) to update the scene graph for adaptive identification of target objects and their spatial relations with respect to the reasoned activity. The DGM guides the model to focus on objects that are critical to the reasoned activity, thereby providing highlighted information for the subsequent route planning based on spatial relations among relevant objects. As illustrated in Figure~\ref{fig:motivation}, our framework can effectively interpret the intention underneath the implicit human instruction, precisely reasoning out the desired activity, and producing detailed and coherent execution steps with inter-step route planning.

The major contributions of this work can be summarized in three aspects. First, we propose a new 3D Reasoning-Driven Planning task based on implicit human instructions, together with a large-scale and high-quality benchmark. The benchmark covers various implicit human instructions and annotations of detailed planning for the intended activity, including executable steps and inter-step routes under the guidance of 3D scene segmentation. The benchmark provides a valuable platform for designs and evaluations in the area of 3D reasoning-driven planning. Second, we design a novel framework that incorporates progressive plan generation for context-aware planning, as well as dynamic graph modulation for capturing critical objects and their relations, achieving effective 3D reasoning-driven planning. Third, extensive experiments demonstrate the superiority and great value of our proposed approach and benchmark in 3D reasoning-driven planning.

\section{Related Work}

\subsection{Embodied Large Language Model Planning}
In the area of embodied task planning, there are various works~\cite{li2025interactive, wu2023embodied, yang2024octopus, yoo2025exploratory, chen2024ll3da, zhu2024unifying, huang2023inner, rajvanshi2024saynav, song2023llm, huang2022language} that leverage large language models to generate executable action plans. Specifically, LLM-based approaches generate action plans either through direct environmental input~\cite{huang2023inner, zhang2024task, rajvanshi2024saynav, song2023llm, li2025interactive, wu2023embodied, chen2024ll3da} or via prompt engineering~\cite{huang2022language}. Recent studies enhance planning by evaluating action affordances~\cite{brohan2023can, hazra2024saycanpay}, employing code-driven policies~\cite{singh2023progprompt}, and leveraging commonsense knowledge from world models~\cite{guan2023leveraging, hao2023reasoning, nottingham2023embodied}. For example, AdaPlanner~\cite{sun2024adaplanner} adaptively refines its self-generated plans using environmental feedback. Grounded 3D-LLM~\cite{chen2024grounded} performs instance segmentation of relevant objects yet lacks route planning capabilities. 
SG3D~\cite{zhang2024task} relies on explicit instructions for sequential object grounding.
Unlike existing approaches~\cite{wu2023embodied,chen2024grounded,zhang2024task} that focus on isolated action steps and rely on explicit user instructions to specify the target activity, our work introduces inter-step route planning and reasons human intentions from implicit user inputs, enabling robots to execute tasks within complex 3D environments.

\begin{table*}[t]

\centering
\setlength{\tabcolsep}{4pt}
\begin{tabular}{l|c|cc|cccc}
\hline
\multirow{2}{*}{Benchmark}  & \multirow{2}{*}{Source} & \multicolumn{2}{c|}{Scale} & \multicolumn{4}{c}{Capabilities of Each IP Pair} \\ \cline{3-8} 
                       &                            &           \#  Scene & \# IP Pair  & \makecell{Task \\ Planning} &  \makecell{Activity \\ Reasoning}  & \makecell{Route \\ Planning}      & \makecell{3D \\Segmentation}              \\ \hline\hline
ALFRED~\cite{shridhar2020alfred}& Simulation  & 120 & 25K &  \checkmark & \texttimes &   \checkmark  &  \texttimes \\
BEHAVIOR-1K~\cite{li2022behavior} &  Simulation & 50  & 1,000  & \checkmark  & \texttimes & \texttimes    &  \checkmark \\
TaPA~\cite{wu2023embodied} & Simulation  & 80 & 15K & \checkmark  & \texttimes &  \texttimes   &  \texttimes \\
SIF~\cite{min2024situated} & Simulation & - & 480 & \checkmark & \texttimes & \checkmark & \texttimes \\
SG3D~\cite{zhang2024task} & Real World  &  4,895 & 22K & \checkmark  & \texttimes &  \texttimes   &  \texttimes \\
SG3D-Nav~\cite{zhang2024task} & Simulation  & 181  & 2,868 & \texttimes  & \texttimes &  \checkmark   & \texttimes  \\
Grounded 3D-LLM~\cite{chen2024grounded} &  Real World & -  & 4.4K  & \checkmark  & \texttimes & \texttimes    & \checkmark  \\
 \hline
ReasonPlan3D (Ours) &  Real World & 1,513 & 27K & \checkmark  & \checkmark &  \checkmark   &  \checkmark \\
\hline
\end{tabular}
\caption{
    Comparison on 3D planning benchmarks.
    The proposed ReasonPlan3D excels in its large-scale and high-quality data samples, enabling a wide spectrum of evaluations, including task planning, activity reasoning from implicit instructions, inter-step route planning, and segmentation. IP Pair denotes the Instruction-Plan pair $\{X_{inst}, Y_{plan}\}$.
    }
\label{tab:comparison_dataset}
\end{table*}

\subsection{Language-Instructed 3D Scene Understanding}
Recent studies in 3D scene understanding~\cite{he2024segpoint, huang2024reason3d, jiang2024multimodal, man2024situational, savva2019habitat} increasingly leverage natural language to enrich contextual knowledge and capture user intentions for human–model interactions. Existing research primarily focuses on tasks such as visual grounding~\cite{huang2022multi, zhu20233d, guo2023viewrefer, yang2024llm, chen2024ll3da, hong20233d, kang2024intent3d}, 3D question answering~\cite{azuma2022scanqa, ma2023sqa3d, parelli2023clip, guo2023point, hong20233d}, 3D referring~\cite{he2024segpoint, huang2021text, qian2024x, wu20243d}, and 3D dense captioning~\cite{chen2023unit3d, chen2023end, hong20233d}. With the advance in large language models (LLMs)~\cite{touvron2023llama}, several recent studies~\cite{wang2023chat, fu2024scene, huang2023embodied, huang2023chat, hong20233d, chen2024ll3da, deng20253d} explore 3D LLMs, aiming to bridge the gap between text and 3D scenes. For example, Chat-Scene~\cite{huang2024chat} employs object identifiers for accurate 3D scene referencing and represents the 3D scene with instance-level features from pre-trained 2D and 3D models. Grounded 3D-LLM~\cite{chen2024grounded} uses referent tokens and contrastive language–scene pre-training to achieve scene-text alignment. Recently, 3DGraphLLM~\cite{zemskova20243dgraphllm} constructs a 3D scene graph to explicitly capture the semantic relations among objects. In contrast, our work introduces a novel 3D Reasoning-Driven Planning task that simultaneously reasons implicit human intentions and performs task planning with integrated inter-step route planning, addressing a critical gap in existing approaches.

\section{ReasonPlan3D Benchmark}

Most existing benchmarks~\cite{wu2023embodied, shridhar2020alfred, chen2024grounded, zhang2024task, hong20233d, rana2023sayplan, li2022behavior, puig2023nopa, puig2021watch, chang2024partnr, min2024situated} for task planning in 3D scenes are not well-suited for exploring the task of 3D Reasoning-Driven Planning. Specifically, existing 3D task planning benchmarks share two major constraints as illustrated in Table~\ref{tab:comparison_dataset}. First, existing 3D task planning benchmarks
are designed for models to follow explicit human instructions regarding the desired activity. They do not provide data for training models for reasoning about humans' implicit intentions about specific activities, thereby limiting the model's capability in discovering the underlying human intentions. Second, most existing 3D task planning benchmarks~\cite{li2022behavior, wu2023embodied, chen2024grounded} focus on isolated action steps while planning a task, neglecting inter-step route planning and hindering the robot assistant from seamlessly interacting with the surrounding 3D environment. We address the two limitations by proposing a data generation pipeline together with a comprehensive benchmark ReasonPlan3D, with more details to be elaborated in the following subsections.

\subsection{Problem Definition}
3D Reasoning-Driven Planning task takes point clouds $P$, multi-view images $I_{mv}$, and user implicit instruction $X_{inst}$ as input, aiming to reason the implicit intention for which activity $A$ the user intend to perform and produce step-by-step plan $\hat{Y}_{plan}$ with inter-step route planning, taking the fine-grained shape and location guidance from segmentation masks $\hat{M}_{ins}$ of relevant objects in $\hat{Y}_{plan}$.

\subsection{ReasonPlan3D Content}
In ReasonPlan3D, each scene represented by point clouds $P$ and multi-view images $I_{mv}$ is paired with multiple triplets $\{X_{inst}, Y_{plan}, M_{3D}\}$. Here, $X_{inst}$ is an implicit human instruction requiring the model to reason the user’s underlying intended activities; $Y_{plan}$ is a step-by-step plan with inter-step route planning; and $M_{3D}$ denotes instance segmentation masks for the objects involved in $Y_{plan}$, providing fine-grained guidance with object shapes and locations for route planning. Specifically, $x_{inst}$ is formulated to avoid explicit statements of the desired activity. For example, instead of saying ``\textit{Please help me make a cup of coffee},'' the instruction would be ``\textit{I want to feel refreshed in the morning in this space. How could you help me?}'' Consequently, $Y_{plan}$ first reasons the intended activity (e.g., preparing coffee) from the implicit instruction, then plans detailed steps and corresponding inter-step route planning in the 3D scene. A sample $Y_{plan}$ in response to the above instruction could be: ``\textit{To help you feel refreshed, the robot assistant will prepare a cup of coffee, with the following steps: Step 1: Walk straight ahead to the kitchen counter. Step 2: Turn 90 degrees left and pick up the water kettle. Fill it with water from the sink. Step 3: Turn 90 degrees right and position the kettle on the stove. Step 4: Move to the coffee machine and add coffee grounds. Step 5: Once the water is heated, pour it into the coffee machine. Step 6: After brewing, grab the mug filled with coffee and deliver it to the table.}"

\begin{figure*}[t]
    \centering
    \includegraphics[width=0.9\linewidth]{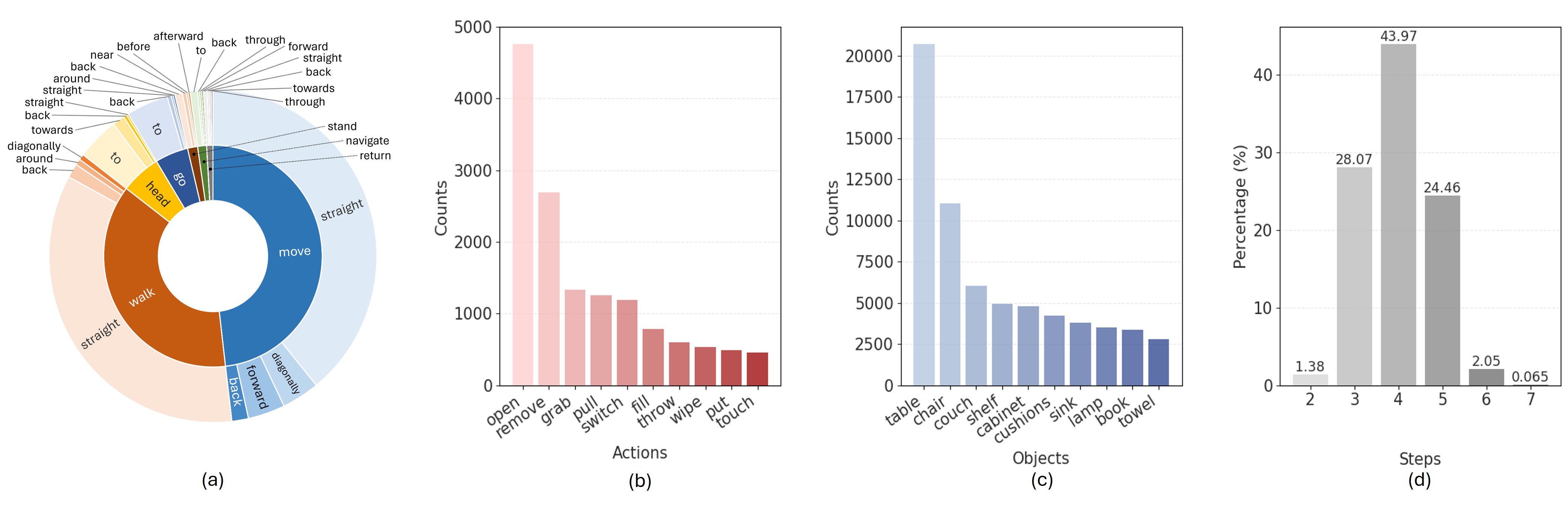}

    \caption{Benchmark statistics. The pie chart in (a) shows the most frequently occurring verbs in inter-step route planning, along with their associated adverbs representing movements. The bar charts in (b) and (c) present actions and their associated objects in the step-by-step plans, and that in (d) show the distribution of answers across different step counts.}
    \label{fig:dataset_statistics}
\end{figure*}

\subsection{ReasonPlan3D Generation Pipeline}

We propose a novel benchmark generation pipeline leveraging GPT-4o to collect data samples. Specifically, we provide GPT-4o with both the scene image and its ground-truth instance segmentation, enabling it to better identify and understand individual objects. The input prompt guides GPT-4o to generate implicit human instructions as well as step-by-step task planning with answers to inter-step route planning. For each generated sample, we conduct human verification and correct errors if any to ensure its quality. We particularly check for errors in route planning and objects, as GPT-4o frequently confuses directions, spatial relations, and objects in 3D scenes. We also check and rectify regarding the feasibility of the generated plans. All rectifications are made strictly based on factual inaccuracies to avoid introducing human preference or subjective bias. 

Overall, GPT-4o generates high-quality, contextually relevant data samples with only about 5\% requiring manual correction, making it a reliable foundation for our benchmark generation.

\subsection{ReasonPlan3D Statistics}

The proposed ReasonPlan3D benchmark comprises a total of 1,513 scenes and 27,608 data samples, including point clouds and multi-view images sourced from ScanNet200~\cite{rozenberszki2022language}. Following the split in \cite{dai2017scannet, rozenberszki2022language}, we divide ReasonPlan3D into training and validation sets with 1201 and 312 scenes, respectively. On average, each scene in the benchmark includes 18.25 instructions, and each textual answer $Y_{plan}$ comprises 3.98 steps and 76.67 words. The benchmark includes 200 distinct object categories for instance segmentation. As illustrated in Figure~\ref{fig:dataset_statistics} (a), ReasonPlan3D encompasses a diverse range of movement categories along with their corresponding adverbs in inter-step route planning. Moreover, Figures~\ref{fig:dataset_statistics} (b) and (c) showcase various actions, such as ``open'', ``remove'', and ``grab'', along with associated objects in these actions, demonstrating the richness of interactions with 3D Scenes. Additionally, as Figure~\ref{fig:dataset_statistics} (d) shows, the distribution of textual answer lengths indicates the complexity of activity planning in our benchmark, with 28.07\%, 43.97\%, and 24.46\% of answers comprising 3, 4, and 5 steps, respectively.

\section{Method}

\subsection{Overall Framework}

\begin{figure*}[t]
    \centering
    \includegraphics[width=0.9\linewidth]{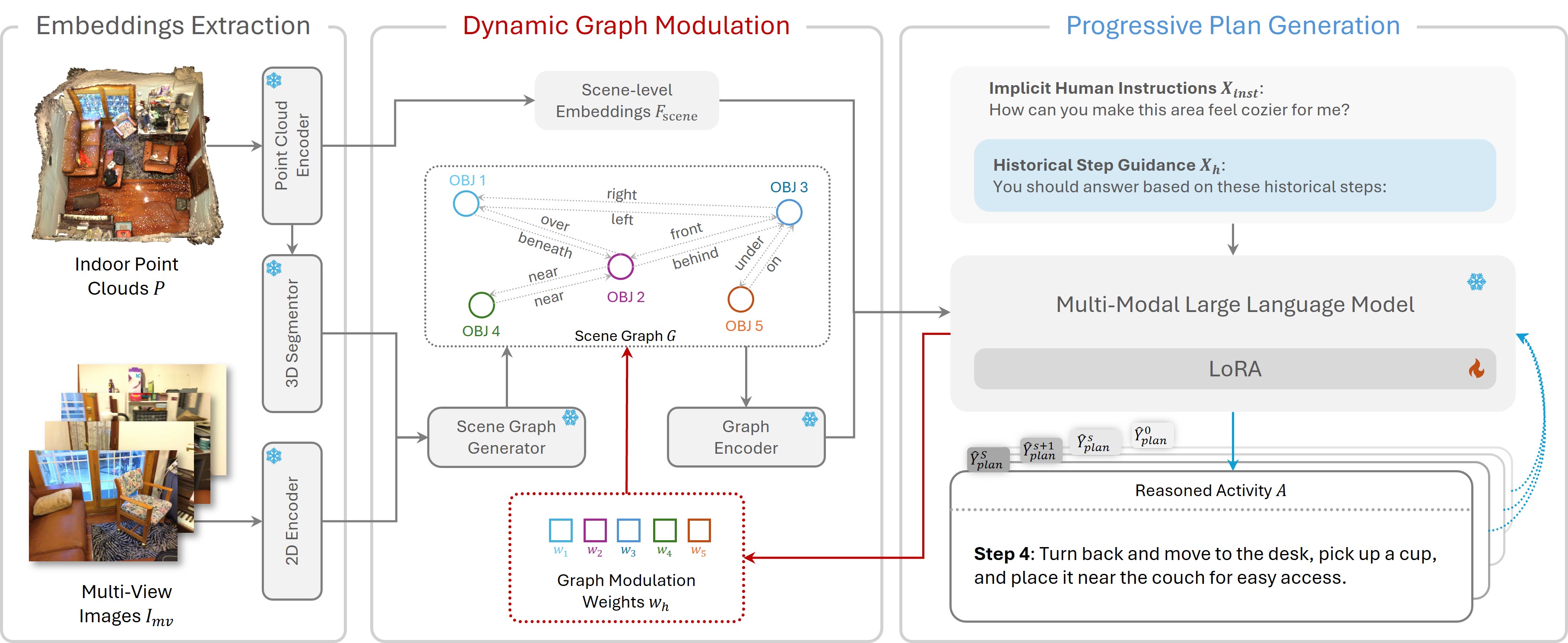}
    \caption{Overall architecture of the proposed SHARP method. Given the point clouds $P$ of a 3D scene, the Point Cloud Encoder generates scene-level embeddings $F_{scene}$, while the 3D segmentor predicts 3D object masks. Besides, the 2D Encoder extracts multi-view image features $F_{mv}$, which are combined with 3D object masks as input to the Scene Graph Generator to obtain scene graph $G$. The generated scene graph $G$ is then fed into the Graph Encoder, together with the scene-level embedding $F_{scene}$ as the inputs of the MLLM. For step $s$, the one-step plan $\hat{Y}_{plan}^{s}$ is generated by referring to previous steps, and the scene graph $G$ is updated by graph modulation weights $w_l$ that emphasize objects and their spatial relations that are critical to the reasoned activity. The snow icon indicates frozen modules, while the fire icon indicates trainable modules.
    }
    \label{fig:overall_architecture}
\end{figure*}

Figure~\ref{fig:overall_architecture} illustrates the proposed framework of \textbf{SHARP}, which enables \textbf{S}cene-grap\textbf{H}-based \textbf{A}ctivity \textbf{R}easoning and \textbf{P}lanning based on implicit user instructions and generates detailed plans with inter-step routes. The input point clouds $P$ are fed into the Point Cloud Encoder and the 3D Segmentor to produce scene-level embeddings $F_{scene}$ and 3D segmentation masks $M_{3D}$, respectively. Besides, the 2D Encoder extracts features $F_{mv}$ from multi-view images $I_{mv}$. Then $M_{3D}$ and $F_{mv}$ are passed to the Scene Graph Generator to obtain the scene graph $G$, which is then processed by the Graph Encoder to obtain object-level embeddings $F_{obj}$. Next, $F_{scene}$ and $F_{obj}$, as well as implicit human instructions $X_{inst}$ are taken as input into the Multi-Modal Large Language Model (MLLM) fine-tuned via LoRA~\cite{hu2021lora}. At step $s$, the MLLM generates a one-step plan $\hat{Y}_{plan}^{s}$ and obtains the graph modulation weights $w_l$. The plans of historical steps $\{\hat{Y}_{plan}^{1}, \cdots, \hat{Y}_{plan}^{s}\}$ are then fed into the MLLM to guide the next step. The graph modulation weights $w_l$ adjust the scene graph $G$ to emphasize objects and their spatial relations relevant to the reasoned activity, benefiting route planning.

\paragraph{Remark 1.} 
To effectively reason about activities with detailed route-aware plans in 3D scenarios, the MLLM is designed to incorporate three key inputs, including the implicit textual instruction, global scene-level embeddings, and 3D scene graph information over reasoned objects.

\subsection{Progressive Plan Generation}

Contextual information is essential for ensuring consistency across steps in our activity planning task. To this end, we propose a progressive plan generation approach that explicitly enforces constraints among different steps by leveraging previously generated steps within a single plan. As illustrated in Figure~\ref{fig:overall_architecture}, for each activity plan, we provide the human's implicit instruction $X_{inst}$ as input to the LLM while generating the first step. For subsequent steps, we prompt the LLM with an input historical step guidance template $X_h$, which incorporates previously generated plans and explicitly instructs the LLM to follow them. Specifically:
\begin{quote}
\ttfamily
Q: \{Human implicit instruction\}. You should answer based on these historical steps: \{plans of historical steps\}.
\end{quote}

Here, \texttt{\{human implicit instruction}\} is replaced with $X_{inst}$ and \texttt{\{plans of historical steps}\} is substituted for $\{\hat{Y}_{plan}^{1}, \cdots, \hat{Y}_{plan}^{s-1}\}$. Besides, for step $s$, only the ground truth of this step's plan $Y_{plan}^{s}$ is provided to supervise the learning of predicting $\hat{Y}_{plan}^{s}$ consistently with historical steps.

To prevent the LLM from generating an excessive number of steps, we introduce a stop token \texttt{[END]} appended to the generated plan for the final step. Once the LLM outputs \texttt{[END]}, it ceases plan generation, keeping the steps count within a reasonable range. The token \texttt{[END]} is incorporated into the LLM’s token vocabulary to enable this functionality.

\subsection{Dynamic Graph Modulation}

We propose a dynamic graph modulation mechanism that adaptively adjusts the importance of nodes and edges in the generated scene graph, thereby enhancing the awareness of relevant objects in the generated action plan. 
Leveraging the one‐step plan semantics generated by the MLLM as a control signal, our proposed mechanism dynamically modulates the scene graph to focus directly on the current action intent in the reasoning-driven planning loop.
Specifically, the generated scene graph $G$ consists of nodes $N$ (representing objects) and edges $E$ (representing spatial relations between two objects). When the generated one-step plan $\hat{Y}_{plan}^{s}$ includes the $i$-th object $N_i$, we emphasize $N_i$, its $K$ neighbors and their spatial relations by applying a larger weight $w_l$ to the embeddings, as formulated below:
\begin{equation}
    \{N_i, N_{ij}, E_{ij}\} * w_l, j\in\{1, \cdots, K\}
    \label{equ:weight_high}
\end{equation}
where node $N_{ij}$ denotes the $j$-th neighbor of $N_i$, and edge $E_{ij}$ represents the spatial relation between $N_{ij}$ and $N_i$.
Here, $K$ neighbors of node $N_i$ are selected by the K-Nearest Neighbors (KNN) algorithm. This mechanism can effectively coordinate with the proposed Progressive Plan Generation approach in the iterative process, as shown in Figure~\ref{fig:overall_architecture}.

\begin{table*}[t]

\centering
\setlength\tabcolsep{2.3pt}
\begin{tabular}{l|c|ccccccc}
\hline
Method       & Venue            & BLEU-1    & BLEU-2 & BLEU-3 & BLEU-4    &   CIDEr  &  METEOR   &  ROUGE  \\ \hline\hline
3D-LLM~\cite{hong20233d}  & NeurIPS 23           &                        34.47                  &     25.73        &     21.59    &  16.52       &    10.02     &  18.61   &   40.29   \\
LL3DA~\cite{chen2024ll3da}  & CVPR 24           &      38.21                                    &     28.92        &  22.74       &  19.86       &     10.93    &  21.55   &   40.72   \\
Chat-Scene~\cite{huang2024chat}  & NeurIPS 24           &      39.47                                    &    31.04         &    \underline{25.80}     &   \underline{21.15}      &    11.67     &   21.94  & 42.28     \\
3DGraphLLM~\cite{zemskova20243dgraphllm}  & arXiv 24           &  \underline{40.29}      &  	\underline{31.37}	     &  25.37     &  20.98     &  	\underline{11.82}     &  	\underline{22.51}	     &  \underline{43.60}
  \\
SHARP (Ours)  & -           &                                    \textbf{52.22}      &       	\textbf{39.30}	     &       \textbf{30.58}	     &       \textbf{24.59}     &       	\textbf{25.51}     &       	\textbf{27.51}	     &       \textbf{43.76}
     \\ \hline
\end{tabular}
\caption{
    Benchmarking on the ReasonPlan3D validation set for the 3D Reasoning-Driven Planning task with evaluation metrics BLEU, CIDEr, METEOR, and ROUGE. Best in bold, second underlined.
}
\label{tab:benchmark_with_sota}
\end{table*}

\begin{table*}[t]

\centering
\setlength\tabcolsep{28pt}
\begin{tabular}{l|ccc}
\hline
Method                   &  BLEU-4 &  CIDEr   &     METEOR  \\ \hline\hline
3D-LLM~\cite{hong20233d}    &  38.02  & 285.46  &  30.13  \\
LL3DA~\cite{chen2024ll3da}  &   42.59    &  302.47   &  33.20 \\
Chat-Scene~\cite{huang2024chat}    &  45.26  &  313.90  &  34.42  \\
3DGraphLLM~\cite{zemskova20243dgraphllm}   &    \underline{47.65}     &   	\underline{329.59}     &   	\underline{36.87}  \\
SHARP (Ours)         & \textbf{50.28}	     &   \textbf{359.53}     &   	\textbf{38.35}
     \\ \hline
\end{tabular}
\caption{
   Benchmarking on the ReasonPlan3D validation set for measuring activity reasoning performance with evaluation metrics BLEU-4, CIDEr, and METEOR. Best in bold, second underlined.
}
\label{tab:activity_reasoning_benchmark}
\end{table*}

By focusing on the included objects, their most relevant neighbors, and corresponding spatial relations, this dynamic graph modulation mechanism improves the awareness of critical objects and their spatial relations which is beneficial for route planning within our task.

\subsection{Training Objective}

For the textual responses generated by the LLM, we optimize the activity reasoning, step-by-step planning, and inter-step route planning using the following loss:
\begin{equation}
    L_{plan} = CE(Y_{plan}, \hat{Y}_{plan}),
    \label{equ:l_plan}
\end{equation}
where $CE(\cdot)$ denotes the cross-entropy loss. We train the proposed model end-to-end using $L_{plan}$.

\section{Experiment}
\subsection{Experimental Settings}

\paragraph{Evaluation Metrics.}
We adopt BLEU~\cite{papineni2002bleu}, ROUGE~\cite{lin2004rouge}, METEOR~\cite{banerjee2005meteor}, and CIDEr~\cite{vedantam2015cider} to evaluate the quality of generated plans for the reasoned activities. In addition, we adopt BLEU-4, CIDEr, and METEOR to measure the performance of activity reasoning. Since our task centers on human-robot interaction and textual response generation, particularly for activity reasoning and planning, we adopt language-based metrics to evaluate the overall quality of the responses.

\paragraph{Implementation Details.}
We conduct experiments on one NVIDIA A40 GPU (48 GB memory), and train our framework for $11k$ iterations with a batch size of $2$ and a total training time of around 9 hours. We employ the AdamW~\cite{loshchilov2019decoupled} optimizer with an initial learning rate of $2e^{-5}$, a weight decay of $0.02$, and a cosine scheduler with $1.1k$ warm-up iterations. We adopt LLaMA-3-8B-Instruct~\cite{llama3modelcard} as our multimodal LLM backbone and apply LoRA~\cite{hu2021lora} with a rank of 16 for efficient fine-tuning. The weight $w_l$ used in the proposed Dynamic Graph Modulation mechanism is set as $2.0$. All experiments are performed on our proposed ReasonPlan3D dataset.

\subsection{Benchmarking with Existing Methods}
Table~\ref{tab:benchmark_with_sota} presents quantitative experiments on the ReasonPlan3D validation set, where all compared methods are trained and evaluated on the same ReasonPlan3D benchmark for fairness. We can observe that the proposed SHARP achieves superior performance across all evaluation metrics, demonstrating its superiority in activity reasoning, task planning, and inter-step route planning. The superior performance is largely attributed to two key factors. First, the proposed Progressive Plan Generation enforces the model to maintain contextual consistency throughout the plan generation process by referring to the plans generated in previous steps. Second, the proposed Dynamic Graph Modulation guides the model toward critical objects and their spatial relations, enhancing 3D scene understanding and boosting route planning accuracy.

In addition, Table~\ref{tab:activity_reasoning_benchmark} presents the activity reasoning performance of SHARP and several state-of-the-art methods over the validation set of ReasonPlan3D. Our approach achieves superior performance across all metrics. Specifically, 3D-LLM~\cite{hong20233d}, LL3DA~\cite{chen2024ll3da}, and Chat-Scene~\cite{huang2024chat} do not employ a scene graph to explicitly guide the model's awareness of critical objects, leading to suboptimal performance when dealing with the activity reasoning task, where comprehending object relations plays an essential role. 
3DGraphLLM~\cite{zemskova20243dgraphllm} employs only scene graphs without scene-level embeddings, leading to inferior 3D scene understanding and weaker performance for activity reasoning.

\begin{figure*}[t]
    \centering
    \includegraphics[width=0.99\linewidth]{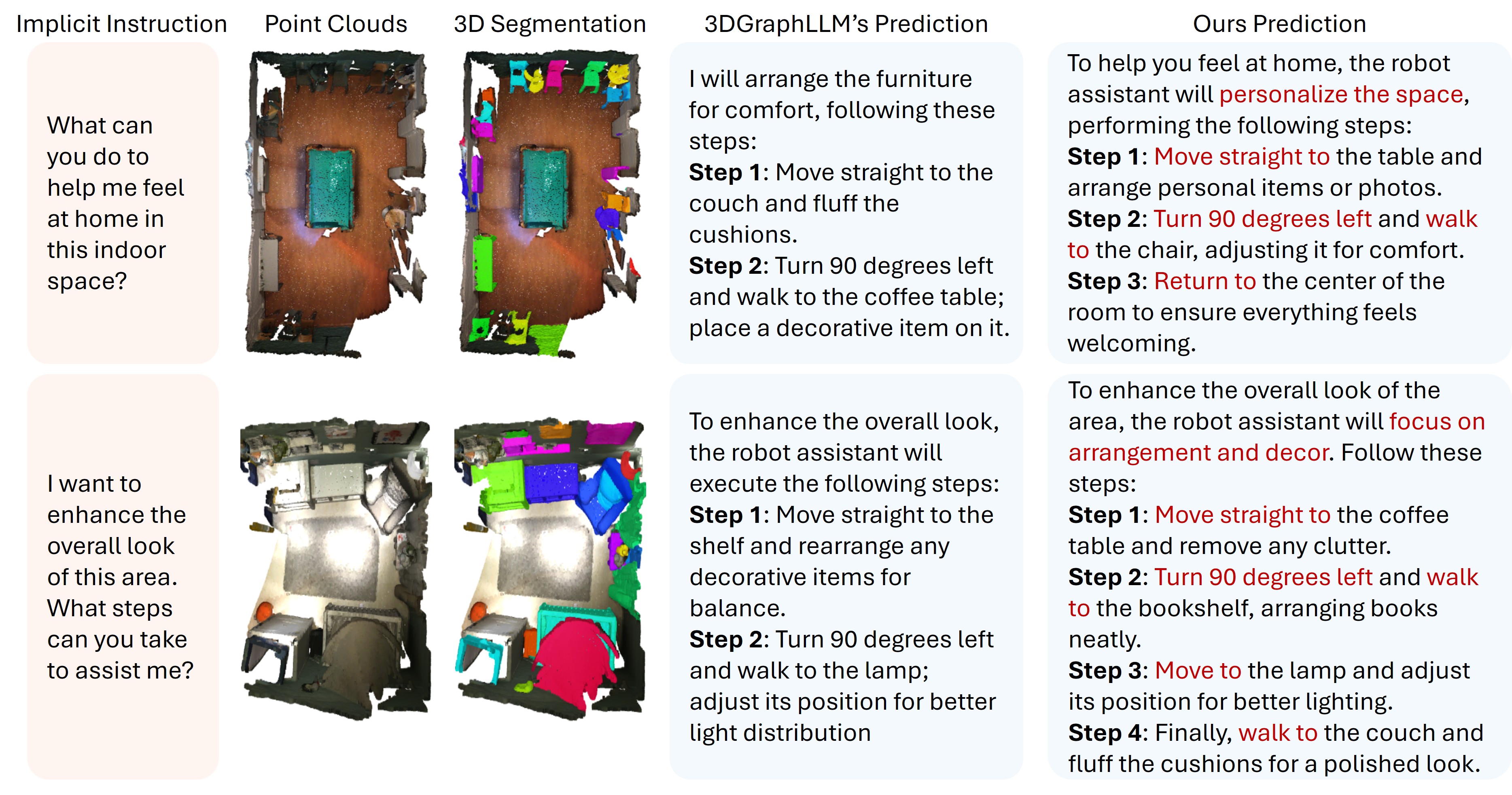}
    \caption{Activity reasoning and planning visualization over the ReasonPlan3D val set. Each example shows an implicit human instruction, the input point clouds of the 3D scene, the 3D segmentation of the scene, and the predictions from 3DGraphLLM and SHARP. Best viewed in color and zoom-in.}
    \label{fig:visualization}
\end{figure*}

\begin{table*}[t]

\centering
\setlength\tabcolsep{10pt}
\begin{tabular}{c|cc|ccccccc}
\hline
Index       &  PPG     &  DGM       & BLEU-1    & BLEU-2 & BLEU-3 & BLEU-4    &  CIDEr  &  METEOR   &  ROUGE     \\ \hline\hline
1*  &            &                                          &    39.61   &      	30.54	   &      24.36	   &      19.87   &      	12.25   &    	21.90   &      	42.03
  \\
2  &   \checkmark        &                                          &  51.46     &    	38.49	     &    29.35     &    	23.82	     &    23.80     &    	27.09     &    	43.39
         \\ 
3  &   \checkmark        &    \checkmark                                       &    \textbf{52.22}      &       	\textbf{39.30}	     &       \textbf{30.58}	     &       \textbf{24.59}     &       	\textbf{25.51}     &       \textbf{27.51}	     &       \textbf{43.76}       \\ \hline
\end{tabular}
\caption{
    Ablation studies over SHARP designs. PPG denotes Progressive Plan Generation, while DGM denotes Dynamic Graph Modulation. The symbol * indicates the baseline. The best results are in bold.
}
\label{tab:ablation_module}
\end{table*}

\paragraph{Qualitative Benchmarking.} Figure~\ref{fig:visualization} provides qualitative results on the validation set of ReasonPlan3D. Each example presents an implicit human instruction, an input point cloud, the corresponding 3D scene segmentation, and the generated textual answers by 3DGraphLLM~\cite{zemskova20243dgraphllm} and the proposed SHARP.
We can observe that SHARP can reason activities from the implicit input instructions accurately while generating high-quality route plans.
In both examples, SHARP surpasses 3DGraphLLM by generating more detailed steps, demonstrating its superior capability in complex planning tasks. Moreover, 3DGraphLLM introduces objects that do not exist in the scene, such as ``couch'', ``cushions'', and ``coffee table'' in the first example, while SHARP exhibits a more precise understanding of the 3D environment, generating plans that accurately reflect the objects present in the scene.

\subsection{Ablation Study}

We conduct ablation studies over the validation set of ReasonPlan3D to evaluate the effectiveness of our designs, including Progressive Plan Generation and Dynamic Graph Modulation in the SHARP framework.

\paragraph{Technical Designs.}
We examine the effectiveness of two key designs in SHARP, namely, Progressive Plan Generation (PPG) and Dynamic Graph Modulation (DGM). As Table~\ref{tab:ablation_module} shows, the baseline (in Row 1) without the two designs does not perform well across all evaluation metrics, largely because it struggles to maintain contextual consistency across steps and fails to capture the objects critical to the reasoned activity. Including the Progressive Plan Generation improves the performance clearly, as the step-by-step plan generation can refer to historical steps and thus preserve contextual information throughout the process. Finally, the best performance is obtained when both designs are included, largely because dynamically highlighting critical objects helps maintain the contextual consistency and allows SHARP to effectively capture implicit intentions and produce coherent and accurate step-by-step route plans.

\section{Conclusion}

This paper presents a novel 3D Reasoning-Driven Planning task that reasons activities from implicit human intentions to satisfy their requirements, decomposes tasks into sequential steps, and plans inter-step routes in complex 3D environments. To this end, we introduce ReasonPlan3D, a large-scale benchmark containing diverse implicit user instructions, step-by-step task planning, inter-step route planning, and fine-grained segmentation annotations. 
On top of this, we propose a novel framework with a progressive plan generation mechanism to ensure that the generated steps remain contextually consistent. Moreover, a scene graph for capturing object relations is constructed, which is dynamically updated to emphasize critical objects and spatial relations for improving route planning accuracy. Extensive experiments confirm the effectiveness of our approach. Future work will investigate more challenging 3D scenarios, further expanding the potential of 3D scene understanding and human-robot interaction in real-world applications.

\bibliography{aaai2025}

\end{document}